%% file: main.tex
\documentclass[10pt,twocolumn,letterpaper]{article}

\usepackage[pagenumbers]{cvpr} 

\input{preamble}

\definecolor{cvprblue}{rgb}{0.21,0.49,0.74}
\usepackage[pagebackref,breaklinks,colorlinks,allcolors=cvprblue]{hyperref}

\def\paperID{4085} 
\def\confName{CVPR}
\def\confYear{2026}

\title{Gated Condition Injection without Multimodal Attention: Towards Controllable Linear-Attention Transformers}

\author{
Yuhe Liu$^{1}$ \quad Zhenxiong Tan$^{1}$ \quad Yujia Hu$^{1}$ \quad Songhua Liu$^{2}$\thanks{Corresponding Author.} \quad Xinchao Wang$^{1}$\footnotemark[1]\\
$^{1}$National University of Singapore \quad 
$^{2}$Shanghai Jiao Tong University\\
{\tt\small \{e1374508, zhenxiong, yujia.hu\}@u.nus.edu \quad liusonghua@sjtu.edu.cn \quad xinchao@nus.edu.sg}
}

\begin{document}
\twocolumn[{%
\maketitle

\vspace{-5mm}

\begin{center}
  \includegraphics[width=0.9\linewidth]{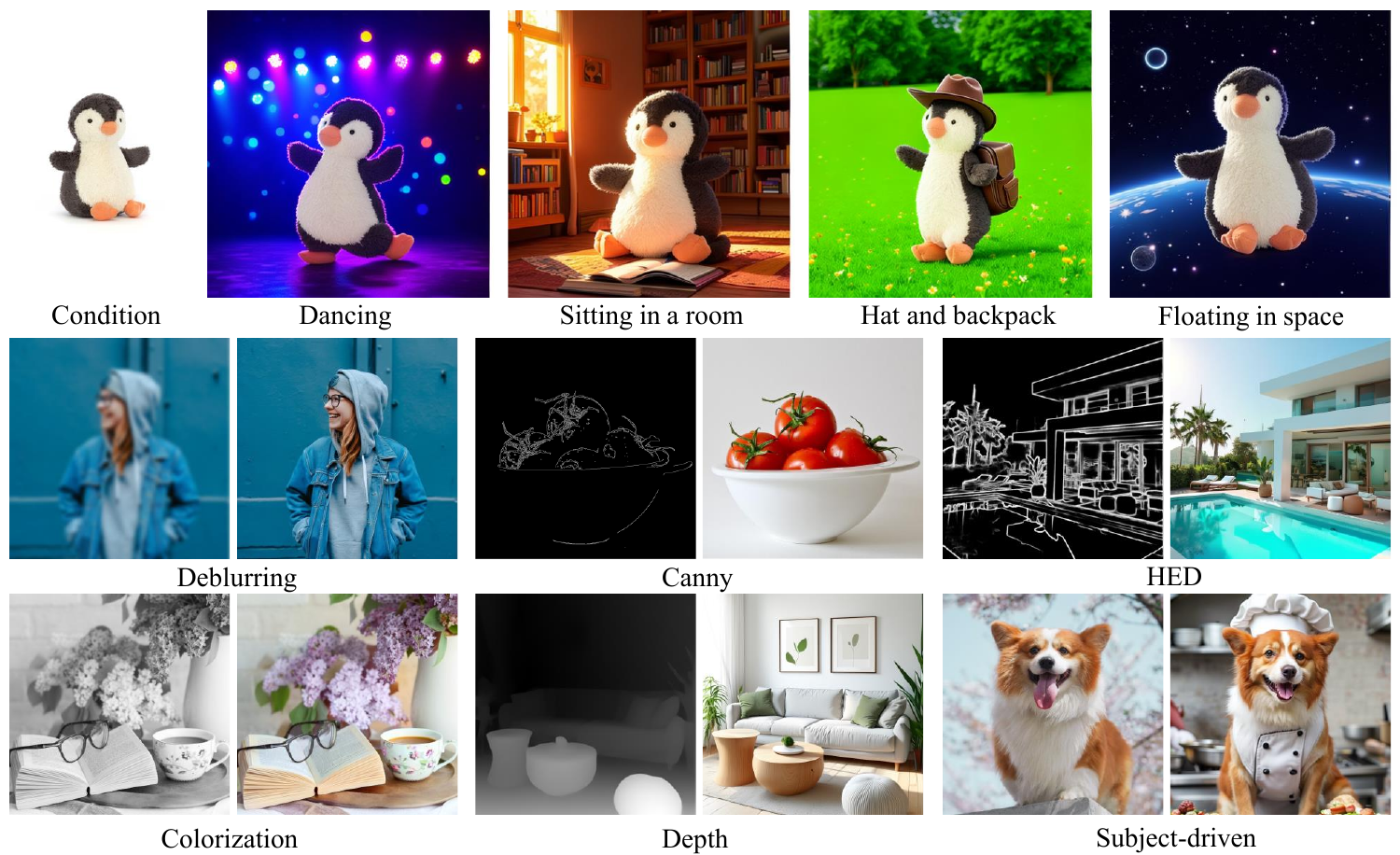}
  \captionof{figure}{Results of our proposed method on SANA-1.0-1.6B for both subject-driven generation and spatially aligned tasks.}
  \label{fig:visual}
\end{center}

\vspace{2mm} 
}]

\begingroup
\renewcommand\thefootnote{}
\footnotetext{$^*$ Corresponding authors}
\endgroup

\input{sec/0_abstract}    
\input{sec/1_intro}
\input{sec/2_related}
\input{sec/3_method}

\input{sec/4_exp}

\input{sec/5_conclusion}
\input{sec/acknowledge}

{
    \small
    \bibliographystyle{ieeenat_fullname}
    \bibliography{main}
}

\input{sec/X_suppl}

\end{document}

%% file: preamble.tex
\usepackage{amsmath,amsfonts,bm}

\usepackage{color}
\usepackage{colortbl}
\usepackage{multirow}
\usepackage{makecell}
\usepackage{wrapfig}
\usepackage{listings}
\usepackage{tabulary}



%% file: sec/0_abstract.tex
\begin{abstract}

Recent advances in diffusion-based controllable visual generation have led to remarkable improvements in image quality. 
However, these powerful models are typically deployed on cloud servers due to their large computational demands, raising serious concerns about user data privacy. 
To enable secure and efficient on-device generation, we explore in this paper controllable diffusion models built upon linear attention architectures, which offer superior scalability and efficiency, even on edge devices. 
Yet, our experiments reveal that existing controllable generation frameworks, such as ControlNet and OminiControl, either lack the flexibility to support multiple heterogeneous condition types or suffer from slow convergence on such linear-attention models. 
To address these limitations, we propose a novel controllable diffusion framework tailored for linear attention backbones like SANA. 
The core of our method lies in a unified gated conditioning module working in a dual-path pipeline, which effectively integrates multi-type conditional inputs, such as spatially aligned and non-aligned cues.
Extensive experiments on multiple tasks and benchmarks demonstrate that our approach achieves state-of-the-art controllable generation performance based on linear-attention models, surpassing existing methods in terms of fidelity and controllability.

\end{abstract}

%% file: sec/1_intro.tex
\section{Introduction}
\label{intro}

Text-to-image generation has undergone a rapid transformation in recent years, largely driven by the remarkable progress of diffusion models~\cite{ho2020denoising, sohl2015deep, song2020score, labs2025flux}. 
Early UNet-based diffusion architectures such as Stable Diffusion~\cite{rombach2021highresolution} established the foundation for high-fidelity image synthesis, demonstrating that iterative denoising can effectively model complex visual distributions. 
As research shifted toward more expressive and scalable architectures, Diffusion Transformers (DiT)~\cite{peebles2023scalable} further pushed the frontier by leveraging large-scale pretraining, global receptive fields, and improved optimization dynamics. 
This evolution from convolutional backbones to fully attention-based designs has led to significant gains in visual quality, semantic alignment, compositional reasoning, and overall generative reliability, positioning diffusion models as the de facto paradigm for state-of-the-art text-to-image pipelines. 

To broaden the applicability of text-to-image models beyond free-form generation, controllable diffusion has emerged as a key direction that injects structural, semantic, or visual priors into the generation process. 
In the UNet era, ControlNet~\cite{controlnet} popularized this paradigm by introducing a trainable copy of the diffusion backbone to encode external conditions such as edges, depth, or poses. These condition features are fused into the main UNet through simple feature addition, enabling effective guidance while preserving the core model’s generative capability. With the transition to DiT architectures, controllability evolved toward more expressive token-level interactions. 
OminiControl~\cite{tan2025ominicontrol} and following works~\cite{zhang2025easycontrol, liu2025step1x} transform conditional inputs into dedicated token sequences and concatenate them with noisy latent tokens, allowing the model to reason over mixed modalities through Multimodal Attention (MM-Attn). This attention-driven formulation greatly enhances flexibility in supporting diverse condition types and provides finer-grained control compared to feature-level fusion in UNet-based designs.

Despite their impressive controllability and generative quality, modern diffusion models are typically deployed on cloud servers due to their substantial computational and memory demands. 
This cloud-dependent paradigm forces users to upload images, sketches, or other personal data for inference, inevitably raising concerns about privacy leakage and limiting the applicability of controllable generation in sensitive or offline scenarios. 
These constraints motivate a shift toward lighter and more deployment-friendly architectures. In particular, linear attention–based diffusion models, such as SANA~\cite{xie2024sana}, offer favorable scalability and significantly reduced memory and compute overhead, making them well suited for on-device execution. In this work, we therefore investigate how to enable controllable diffusion generation within these efficient linear-attention backbones, aiming to achieve both strong control capability and secure, private, on-device generation. 

To this end, we begin by examining ControlNet and OminiControl within linear-attention diffusion models. Unfortunately, both attempts fail to yield satisfactory results. ControlNet’s feature-fusion mechanism implicitly assumes spatial alignment between the input condition and the latent representation. While this works well for depth, edge, or pose maps, it breaks down in spatially unaligned scenarios such as subject-driven generation, where object shapes, poses, or layouts often undergo significant geometric variations. On the other hand, although OminiControl offers greater flexibility by encoding conditions as token sequences and interacting with noisy latents through Multimodal Attention, 
this mechanism proves unsatisfactory under linear attention.
And this method converges slowly on spatially aligned tasks.
These observations highlight the need for a more flexible and efficient control design.

To address these issues, inspired by the concept of attention sinks~\cite{qiu2025gated}, we design a new gating mechanism that enables more flexible and efficient controllability. Specifically, we introduce a learnable gate module that selectively fuses token information
that would otherwise be suppressed by linear-attention interactions. Rather than relying on full attention like MM-Attn, the gate adaptively determines which conditional cues should be preserved and how strongly they should influence each layer, effectively compensating for the information compression introduced by linear attention. This gated interaction not only accelerates training but also enhances conditional feature extraction, thereby improving overall model performance, forming the foundation of our unified controllable generation framework.

\begin{figure}
    \centering
    \includegraphics[width=0.999\linewidth]{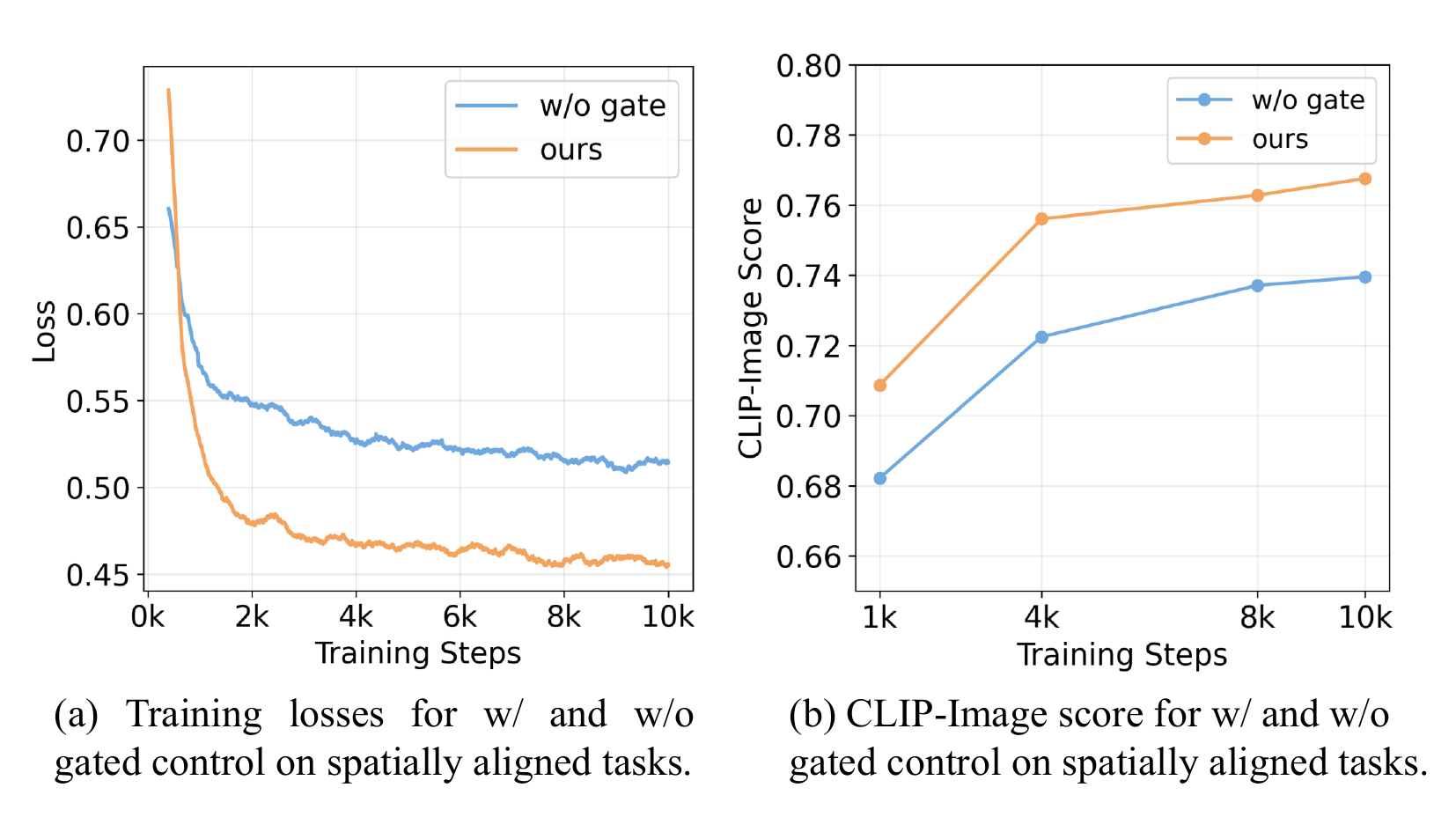}
    \vspace{-0.8cm}
    \caption{Convergence behavior. The introduction of gated modulation results in a substantially steeper decline in training loss compared with interaction mechanisms that rely solely on attention, indicating that the model learns conditional information more rapidly and more effectively, ultimately achieving a lower loss. This trend is also reflected in the CLIP-Image scores, where w/ gated method consistently outperforms the baseline from the earliest training stage and maintains a clear lead throughout.}
    \vspace{-0.3cm}
    \label{fig:curve}
\end{figure}

Across a diverse set of controllable generation tasks, our framework demonstrates substantial benefits in both efficiency and generalization. On spatially aligned conditions such as edge maps and depth maps, our method achieves faster convergence compared to prior approaches, while maintaining the same or better output fidelity. Notably, this acceleration does not come at the expense of flexibility: on spatially unaligned tasks such as subject-driven generation, where object identity, pose, and geometry may shift significantly, our gating mechanism preserves stable convergence and high-quality controllability. The entire improvement requires only 0.09M additional parameters, an overhead that is effectively negligible relative to modern DiT-based diffusion models. Compared with ControlNet, our approach extends far beyond spatial alignment, enabling robust control over non-spatial semantics. 
Compared with OminiControl, it achieves comparable or superior controllability while significantly accelerating convergence on spatially aligned tasks, making it particularly suitable for linear-attention backbones and on-device deployment.

Our contributions can be summarized as follows:
\begin{itemize}
    \item To the best of our knowledge, we present the first controllable generation framework specifically tailored for linear-attention diffusion backbones, enabling efficient on-device deployment while preserving conditional information through gated conditioning.
    \item We introduce a learnable gating module that mitigates the information loss inherent in linear-attention interactions, accelerates convergence, and supports expressive cross-modal conditioning.
    \item The proposed method is highly versatile and effective. Extensive experiments demonstrate that our method achieves strong controllability across a wide range of tasks, including both spatially aligned conditions (e.g., edges, depth) and spatially unaligned scenarios (e.g., subject-driven generation).
\end{itemize}

%% file: sec/2_related.tex
\section{Related work}

\subsection{Generative models}
Recent advances in generative models, including GANs~\cite{gans, goodfellow2014generative, karras2019style}, VAEs~\cite{kingma2013auto}, and diffusion-based methods~\cite{ho2020denoising, sohl2015deep, song2020score}, have enabled high-fidelity image synthesis. Transformer-based architectures~\cite{peebles2023scalable} further improve conditional generation by modeling complex dependencies.
Despite these successes, existing models remain constrained by heavy computation and inflexible controllability~\cite{controlnet, ipadapter, mou2024t2i}.
Current research therefore emphasizes efficiency-aware diffusion sampling~\cite{zhou2024simple, salimans2022progressive, zhou2025few} and hybrid autoregressive–diffusion approaches~\cite{xie2024show, deng2024causal} that balance quality with computational practicality.

\subsection{Conditional control}
Conditional generation seeks to guide the synthesis process with external signals such as text, masks, poses, or semantic layouts. Early approaches~\cite{wang2018high, perarnau2016invertible, sricharan2017semi, mishra2018generative, won2022physics}
concatenate conditioning vectors into the latent representation, while modern diffusion-based methods~\cite{controlnet, ipadapter} employ cross-attention or adapter pathways to inject control at multiple feature levels. These techniques enhance alignment and fine-grained consistency but often incur high training and inference costs. Recent studies~\cite{customdiffusion, tan2025ominicontrol} explore dynamic fusion modules, and unified conditioning frameworks to achieve general condition control, which motivate our design of a more adaptive and much more efficient conditioning mechanism.

\subsection{Gated mechanism}
Gating mechanisms were first developed in recurrent networks such as LSTM~\cite{graves2012long} and GRU~\cite{chung2014empirical} to regulate information flow and stabilize long-term dependencies. Subsequent works extend this principle to feed-forward and attention architectures through gated linear units and conditional routing in mixture-of-experts (MoE) models~\cite{dao2024transformers, gu2024mamba, lepikhin2020gshard}. 
In generative modeling, gating can be useful for selective feature fusion and improving compositional coherence. By dynamically modulating inter-pathway communication, gated designs can enhance both controllability and efficiency. 
Building on these insights, we propose our gated control method.

%% file: sec/3_method.tex
\section{Methodology}
\label{method}

\begin{figure*}
    \centering
    \includegraphics[width=0.999\linewidth]{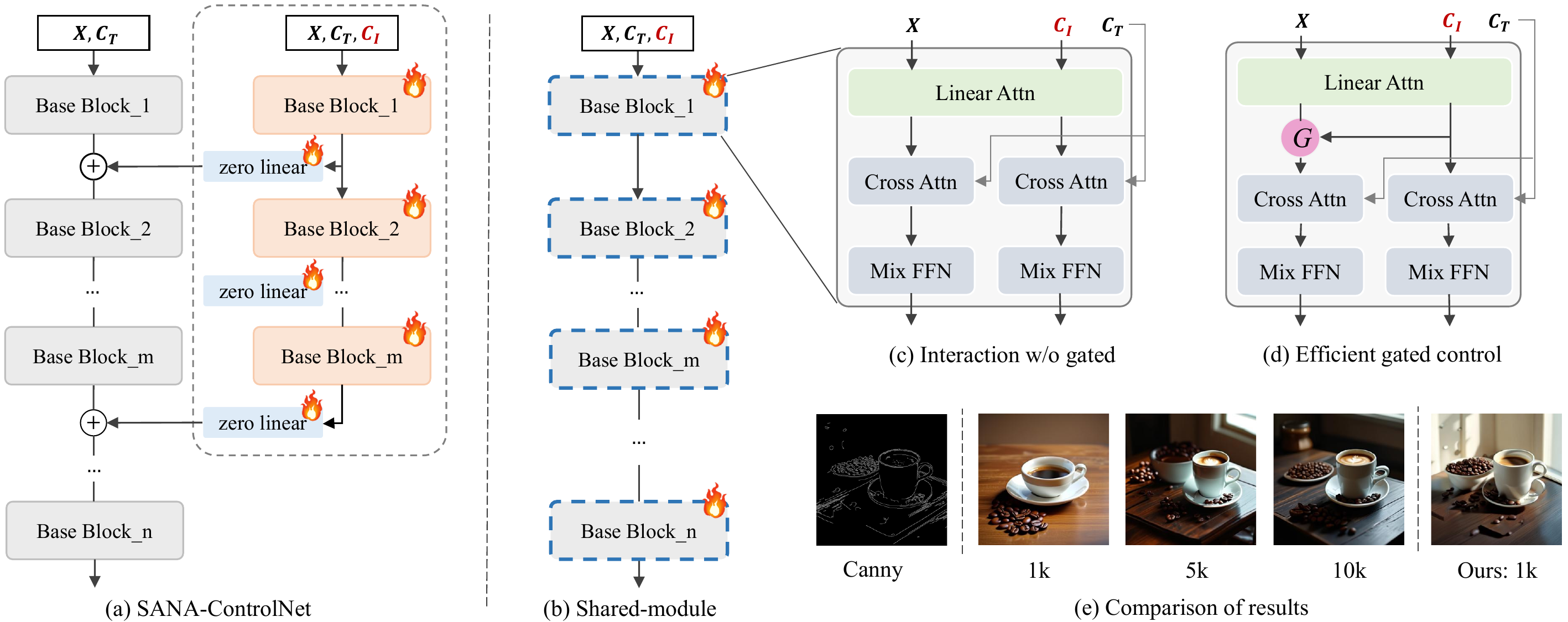}
    \caption{Exploration of different methods and comparison of results. We adopt a shared module to handle the noisy latent and image condition, thus maximally retaining the original information flow of the model. The internal interaction enables flexible and general conditional control. Furthermore, a unified gating mechanism allows both non-spatial and spatial information to be effectively injected, improving performance while greatly accelerating the convergence of spatial tasks.}
    \vspace{-0.4cm}
    \label{fig:model}
\end{figure*}

\subsection{Preliminary}
\noindent\textbf{Flow matching}
In flow-based frameworks~\cite{lipman2022flow, liu2022flow, xie2024sana}, the denoising process is formulated as a probability density flow, with the vector field $u_t(x)$ learned by a neural network:
\begin{equation}
\mathcal {L}_{F M}:=\mathbb{E}_{t, p_t(x)}\left[\left\|v_t(x) - u_t(x)\right\|_2^2\right]
\label{eq3}
\end{equation}
where $p_t(x)$ represents the probability density path, $x \sim p_t(x)$ and $t \sim \mathbb{U}[0,1]$. In this paper, we conduct our experiments primarily based on SANA~\cite{xie2024sana}, which is an efficient generative model based on rectified flow.

\textbf{Attention mechanism}
In the attention module, the input $X $ is first transformed into queries $Q$, keys $K$ and values $V$ using projection matrices $W_Q, W_K$ and $W_V$:
\begin{equation}
    Q= XW_Q, ~K= XW_K, ~V= XW_V,
\end{equation}
where $W_Q, W_K, W_V \in \mathbb{R}^{d_{\text{model}}\times d}$ and $Q, K, V \in \mathbb{R}^{n\times d}$.

For standard multi-head softmax attention, the output is a weighted sum of values, where the weights are computed from the dot products between queries and keys followed by softmax normalization, as:
\begin{equation}
\text{Attention}(Q, K, V) = \text{Softmax}(\frac{(QK^{\top})}{\sqrt{d}})V.
\end{equation}

But it incurs a
computational cost that scales quadratically with the input length, posing a significant bottleneck
when handling long sequences.
To alleviate this, linear attention~\cite{katharopoulos2020transformers, xie2024sana} replaces the exponential dot product with a kernel decomposition, such that:
\begin{equation}
\text{Attention}(Q, K, V) = \phi(Q)~(\phi(K)^T~V).
\end{equation}

This can reduce computational complexity, especially when generating images of larger sizes. SANA~\cite{xie2024sana} adopts ReLU-based linear attention~\cite{katharopoulos2020transformers}.

\subsection{GateControl}
We aim to design a \textbf{general}, \textbf{minimal} and \textbf{efficient} controllable generation framework that accepts flexible control condition inputs while achieving \textbf{alignment} and \textbf{rapid convergence}. 
Motivated by this, we present GateControl, which we will describe in the following.

\subsubsection{Shared-module encoding}

Prior studies~\cite{controlnet, ipadapter, mou2024t2i} have focused on incorporating image condition information. 
ControlNet~\cite{controlnet} employs a trainable copy of model blocks to process the conditions and injects the features through direct addition, as:
\begin{equation}
    h_x \leftarrow h_x + h_c.
\end{equation}

However, after implementing this approach in Figure~\ref{fig:model}(a), we observed two major issues:
(1) it introduces additional architectural components with excessive parameters, which degrades training and convergence efficiency; and
(2) it exhibits insufficient adaptability when facing non-spatially aligned inputs, particularly in subject-driven cases.

To reduce model parameters, we employ a shared-module
strategy, inspired by~\cite{tan2025ominicontrol}.
As shown in Figure~\ref{fig:model}(b), both the image condition and the noisy latents are transformed through the same VAE encoder into a shared parameter space, followed by processing through an identical model structure.
Heavy adapters typically rely on separate encoders (e.g., CLIP), which require additional alignment. By contrast, our method uses a shared backbone, naturally reducing extra parameters.
Furthermore, we apply LoRA fine-tuning to adapt the model to the new conditional inputs, effectively avoiding the expense of full fine-tuning.

As illustrated in Table~\ref{table:param}, our parameter reuse strategy eliminates the need for an extra trainable model copy, image processor, or cross-attention interaction module used in ControlNet and IP-Adapter.
This design retains the original model architecture and significantly reduces the number of trainable parameters to 18.9M (less than 1.18\%), compared to 590M additional parameters required by ControlNet.

\begin{table}[t]

    \scriptsize
    \centering
    \begin{tabular}{l|c|c|c}
        \toprule {Methods}                   & {Base model}                                               & {Parameters}                                            & {Ratio}                                                             \\
        \midrule \midrule ControlNet         & \multirow{3}{*}{\makecell{SD1.5 \textcolor{gray}{/} 860M}} & 361M                                                    & $\sim$42\%                                                          \\
        T2I-Adapter                          &                                                            & 77M                                                     & $\sim$9.0\%                                                         \\
        IP-Adapter                           &                                                            & 449M                                                    & $\sim$52.2\%                                                        \\
        \midrule 
        ControlNet & \multirow{2}{*}{\makecell{SANA \textcolor{gray}{/} 1.6B}} & 590M & $\sim$36.9\% 
        \\
        IP-Adapter &  & 33.7M & $\sim$2.11\% 
        \\
        \midrule \rowcolor{cvprblue!15} \makecell{Lora interaction \\ + Ours}  &  \makecell{SANA \textcolor{gray}{/} 1.6B}        & \makecell{18.9M  \\ + 0.09M w/ Gate} & \makecell{$\sim$1.18\% \\ + $\sim$0.006\% } 
        \\
        \bottomrule
    \end{tabular}
    \vspace{-1mm}
    \caption{Additional parameters introduced by different image conditioning methods.
For our method, results are reported with LoRA rank set to 16.}
    \label{table:param}
    \vspace{-1em}
\end{table}

\subsubsection{Internal interaction}

To achieve flexible and general conditional control in linear attention models, we design the internal interaction mechanisms of the model. 
As depicted in Figure~\ref{fig:model}(b), 
we treat the tokens of the noisy latent $X$, text conditions $C_T$ and image conditions $C_I$ as a whole to get $[X;C_T;C_I]$, 
enabling every block to process them together as an integrated input.
In our paper, we instantiate an OminiControl-style~\cite{tan2025ominicontrol} design on the SANA as a baseline~(in Figure~3(c) ``w/o gate''), which has the same backbone, training data and schedules as our full method.
The core mechanism of OminiControl is attention-based control.
The unified representation allows the model to perform bidirectional attention interaction in the linear attention module, thereby facilitating flexible information fusion between $X$ and $C_I$.
Meanwhile, 
both latent and condition tokens maintain independent interactions with the text condition via cross-attention and perform fusion within the Mix-FFN.

This design maximally retains the original information flow of the model, allowing it to adapt to the newly introduced conditions with minimal modifications.
In addition, such internal interaction makes flexible and general conditional control possible. 
Our experiments confirm that this \textbf{bidirectional linear attention} is sufficient for condition injection, supporting both spatial control tasks (\eg depth, Canny, coloring, deblurring) and non-spatially aligned tasks such as subject-driven generation~(see in Section~\ref{exp}).

In practice, however, this design also presents a drawback: \textbf{without explicit spatial information} injection or alignment, the model learns and converges much more slowly on spatially aligned tasks, greatly diminishing its effectiveness in spatial control scenarios.
This phenomenon is also observed in OminiControl~\cite{tan2025ominicontrol}. Due to the stronger requirement for spatial alignment, spatial tasks require significantly more training iterations (50k vs. 15k) than subject-driven tasks, as reported in the paper.

\subsubsection{Efficient gated control}
To alleviate the above issues, we 
draw inspiration from the gating mechanisms, which 
can selectively filter tokens based on their importance. 
Recent research~\cite{qiu2025gated} also further reveals that gating can help alleviate the attention sink issue in large language models~(LLMs).

Inspired by these observations, we design an adaptive gated control token-fusion mechanism.
As illustrated in Figure~\ref{fig:model}(d), a gated control is applied after the linear attention layer, allowing the information from the noisy latent and the image condition to be adaptively fused.
We observe that even a \textbf{minimal token-wise gating} mechanism could effectively filter tokens by strength, enabling adaptive token fusion with negligible cost.

\begin{figure}
    \centering
    \includegraphics[width=0.96\linewidth]{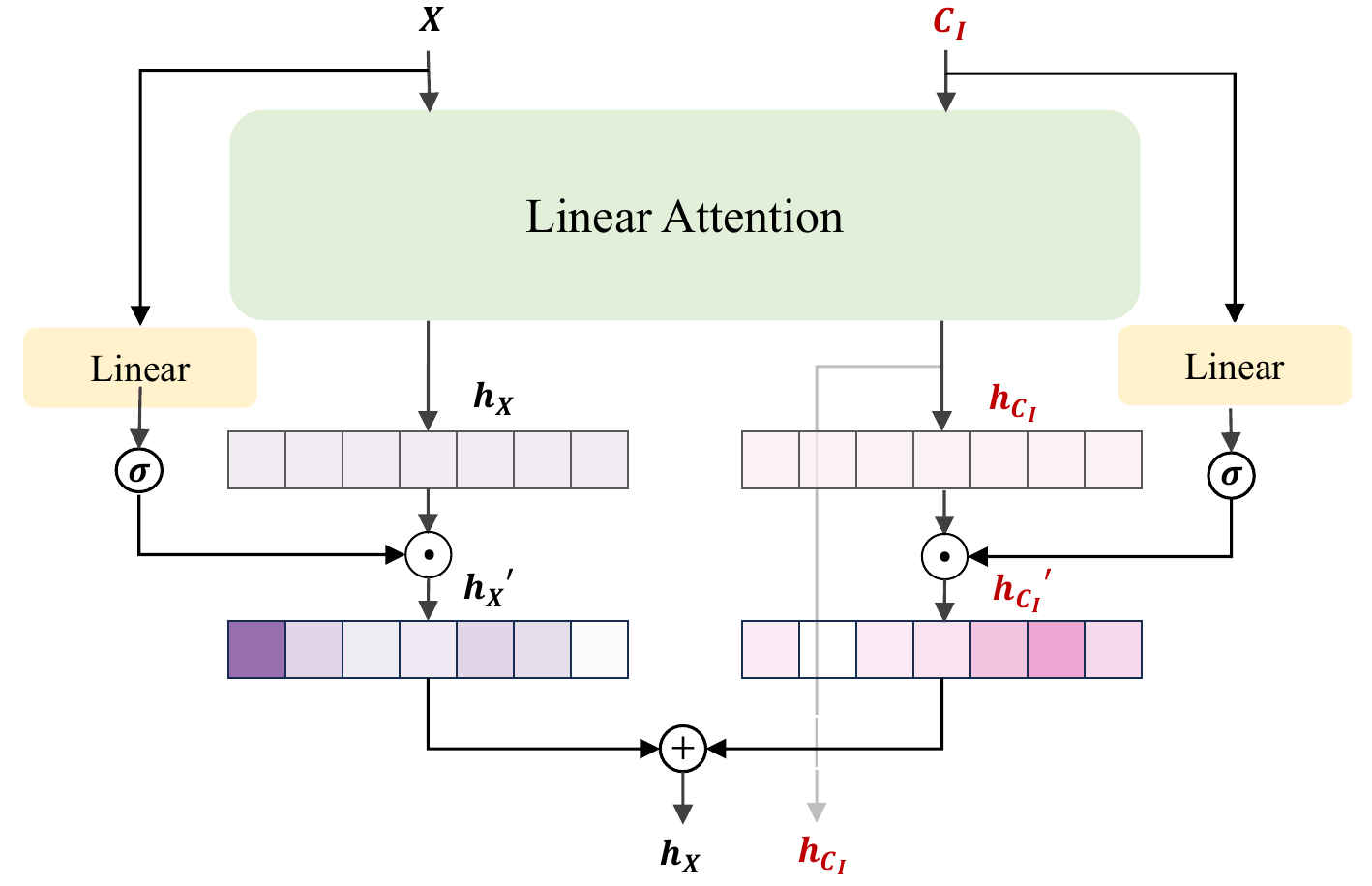}
    \vspace{-0.3cm}
    \caption{Details of the gated control method. Even a minimal token-wise gating mechanism could effectively filter tokens by strength, enabling adaptive token fusion with negligible cost.}
    \vspace{-0.6cm}
    \label{fig:gated}
\end{figure}

We detail our gated control design in Figure~~\ref{fig:gated}, in which we adopt data-dependent token strength filtering in the fusion.
For hidden states $h_X$, we perform gated modulation to enable per-token information control, as:
\begin{equation}
    {h_X}^{\prime}= g(h_X, X, W_{g1}, \sigma) =  \sigma(XW_{g1}) \odot h_X ,
\end{equation}
where $h_X$ is the tokens to be modulated~\footnote{We adopt the feature after linear attention as the input, see the ablation in Section~\ref{exp}.}, and X is another input used to compute the gating scores.
We compute the gating score by applying a linear transformation followed by a sigmoid activation.
$W_{g1}$ denotes the learnable parameters of the gate, and the sigmoid function $\sigma$ squashes the linear mapping into the $[0, 1]$ range, producing a gating score that acts as a soft, dynamic filter and can selectively preserve or suppress features.

Similarly, for the hidden states of image condition $h_{C_I}$:
\begin{equation}
{h_{C_I}}^{\prime} = g(h_{C_I}, C_I, W_{g2}, \sigma) =  \sigma(C_IW_{g2}) \odot h_{C_I}.
\end{equation}

Note that the gating scores are computed independently for each token. This allows each token to decide whether its information should be retained or aggregated by itself, without relying on interactions with other tokens.
After gated modulation, the tokens from hidden states and image hidden states are added for fusion:
\begin{equation}
    h_X \leftarrow {h_X}^{\prime} + {h_{C_I}}^{\prime}.
\end{equation}

In this paper, we thoroughly investigate the mechanism of gated control for information integration. Specifically, we explore four key aspects:
(1) \textbf{Whether to use gating.} We analyze the impact of applying the gated mechanism;
(2) \textbf{Positions.}
Three possible positions are examined: 
(a) after the self-attention layer, (b) after the cross-attention layer, and (c) after the Mix-FFN layer;
(3) \textbf{Token-wise, element-wise or direct addition.} Considering the dimensional properties of each token, we further explore: (a) Token-wise gating: each token is assigned an individual gated score; (b) Element-wise gating: each element within a token has its own gated score; (c) Add directly: the image features are directly added to the noisy latent;
(4) \textbf{Input features.} We consider two types of input features, (a) those taken before self-attention and (b) those taken after self-attention.

Our findings indicate that token-wise gating using features before self-attention to compute gating scores achieves the best and most stable results. Full ablation details are reported in Section~\ref{exp}.
Unless stated otherwise, this configuration is used as our default gated control strategy.
This design also has the advantage of introducing an extremely small number of additional parameters. As shown in Table~\ref{table:param}, the proposed gated control adds only $0.09$M parameters, accounting for merely $0.006\%$ of the original parameters of SANA.

With explicit spatial alignment, it achieves not only superior performance but also over $10\times$ faster convergence on spatial tasks.
As shown in Figure~\ref{fig:model}(e), in the canny-to-image task, with only 1k training iterations, our gated fusion model already outperforms the 10k steps baseline without gated spatial alignment. 
Furthermore, the proposed design is general and can accommodate both spatially aligned and subject-driven tasks.

%% file: sec/4_exp.tex
\section{Experiments}
\label{exp}

\input{tables/main_table}

\subsection{Experimental setups}
\noindent\textbf{Model and tasks.}
We build our method upon SANA~\cite{xie2024sana}, which is an efficient text-to-image
model using linear attention.
Experiments are conducted on spatially aligned tasks (\eg Canny-to-image, depth-to-image, colorization, deblurring, and HED-to-Image) and subject-driven generation.
By default, SANA-1.0 is used as the base model.

\noindent\textbf{Training settings.}
We adopt LoRA finetuning for the entire model, with a default rank of 16.
The model is optimized using the Prodigy optimizer, with both safeguard warmup and bias correction enabled. The weight decay is set to 0.01. We adopt an initial learning rate of 1.
Experiments are conducted on 4 NVIDIA H200 GPUs with a per-GPU batch size of 16.
For subject-driven generation, we train on the 1024×1024 subset of Subject200K for 20K steps.
While for spatially aligned tasks, we finetune on 10K images from Text-to-Image-2M for 10K steps.

\noindent\textbf{Baselines.}
We compare our method with several representative baselines, including ControlNet and T2I-Adapter built upon Stable Diffusion 1.5, as well as the ControlNet and OminiControl variant implemented on SANA. For subject-driven generation tasks, we further compare our approach with the IP-Adapter method.

\noindent\textbf{Evaluation on spatially aligned tasks.}
We conduct evaluations and comparisons on both spatial-aligned and subject-driven tasks.
For spatial-aligned tasks, we use the COCO 2017 validation set containing 5,000 images for generation and metric computation.
Corresponding conditional inputs (e.g., edges, depth maps, etc.) are generated from them to serve as model conditions.
For all generations, we fix the random seed to 32 to ensure reproducibility.
We evaluate our model from two perspectives: controllability and image quality.
To assess image quality, we employ multiple metrics, including FID, SSIM, MUSIQ, and CLIP-image and text similarity scores.
Following OminiControl, we measure controllability by computing the F1 score between the edge maps of the generated images and those of the input conditions.
For tasks such as depth-to-image, we instead compute the MSE between the depth maps extracted from the generated and reference images.

\noindent\textbf{Evaluation on subject-driven tasks.}
For subject-driven generation tasks, we evaluate our model using the DreamBooth dataset, which contains 30 subjects and 25 prompts per subject for image synthesis.

\subsection{Main results}

\noindent\textbf{Quantitative comparison.}
Table~\ref{tab:main_results} compares our method with baseline approaches across five spatially aligned tasks. We evaluate controllability (F1 for Canny and MSE for the remaining tasks), image quality (SSIM, FID, and MUSIQ), and alignment (CLIP-Text and CLIP-Image). Following the original settings, results on SD-1.5 are measured at a resolution of 
$512\times512$. As shown, our approach generalizes effectively across all spatial alignment tasks and consistently demonstrates clear advantages at different resolutions. 
Our method achieves comparable or even superior performance relative to SD-1.5 based conditional control techniques. When compared with ControlNet and OminiControl results on SANA, our method delivers overwhelming improvements: for instance, in the colorization task, it reduces FID from 24.95 to 10.28, and in the HED task, it cuts MSE from 2320 to 1168, which is an improvement of more than 50\%. Across colorization, HED, and deblurring, our method achieves substantial gains in FID, MUSIQ, and CLIP-Image scores, while the F1 and MSE metrics further highlight its stronger controllability and more precise adherence to spatial constraints.

\vspace{1mm}
\noindent\textbf{Qualitative comparison.}
Figure~\ref{fig:visual} illustrates the effectiveness of our approach on both spatial alignment and subject-driven tasks. As shown, our method produces high-quality results in deblurring, colorization, and various structure-to-image settings, including Canny-to-image, depth-to-image, and HED-to-image, achieving rich color expression, coherent scene composition, and visually satisfying outputs. For subject-driven generation, our approach also demonstrates strong generalization: it adapts flexibly to diverse scenarios such as dancing, sitting, and floating, and integrates seamlessly with environmental factors like lighting conditions, indoor–outdoor settings, and sunlight. Moreover, our method maintains consistent subject identity while enabling precise and localized modifications, such as adding accessories like hats or backpacks, which highlights its strong controllability and fine-grained editing capability.

\begin{figure*}
    \centering
    \includegraphics[width=0.75\linewidth]{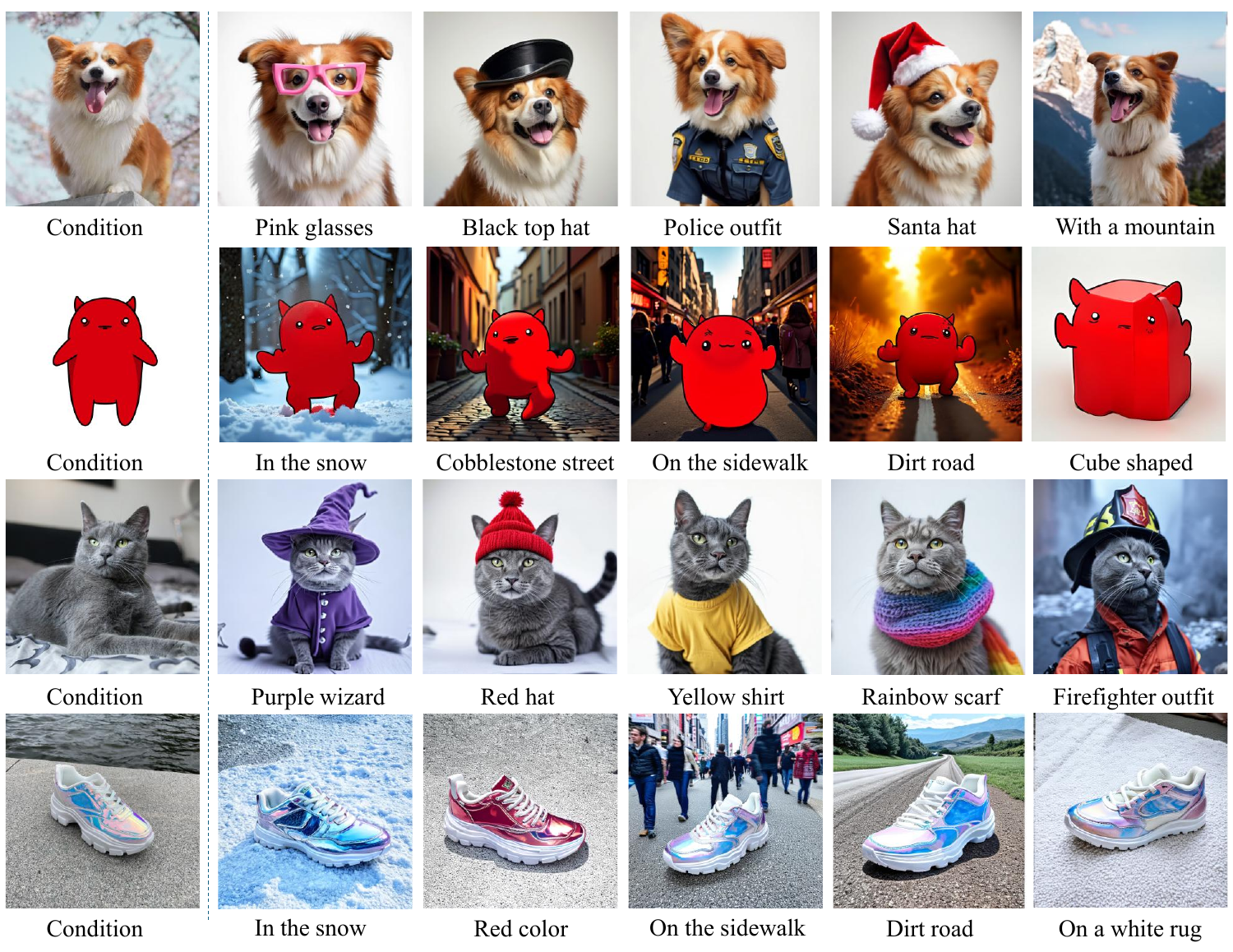}
    \vspace{-0.2cm}
    \caption{Visualization results of our model on subject-driven tasks. The conditioning images are sourced from the DreamBooth dataset.}
    \vspace{-0.4cm}
    \label{fig:subject}
\end{figure*}

Figure~\ref{fig:subject} presents additional results on subject-driven generation, demonstrating the flexibility and precision of our method in preserving subject identity while enabling diverse and fine-grained modifications. Our approach accurately captures the distinctive characteristics of the provided subject and supports a wide range of controlled edits, such as dressing a cat in a firefighter uniform or a wizard outfit, adding hats or accessories, or altering the color of shoes with ease. In the cartoon character examples, the model maintains core identity features while convincingly adapting to different scene contexts, personalities, poses, and expressions, and can even transform the character into geometric shapes such as cubes without losing identity consistency. These results suggest that our gated mechanism holds strong potential for applications in personalized character creation and video generation, where maintaining subject identity while naturally embedding it into varied environments is crucial.
We leave this for future exploration.
Further visual comparisons with other methods can be found in the appendix.

\noindent\textbf{Convergence behavior.} 
Beyond qualitative improvements, our method also exhibits clear advantages in convergence behavior. As shown in Figure~\ref{fig:curve}, the introduction of gated modulation results in a substantially steeper decline in training loss compared with interaction mechanisms that rely solely on attention, indicating that the model learns conditional information more rapidly and more effectively, ultimately achieving a lower loss. Figure~\ref{fig:model}(e) further highlights this advantage: whereas the baseline requires more than 10k steps to gradually acquire spatial correspondence, our method captures explicit positional relationships within the first 1k steps, demonstrating significantly more efficient learning of spatial structure. This trend is also reflected in the CLIP-Image scores, where our method consistently outperforms the baseline from the earliest training stage and maintains a clear lead throughout. The absence of gated information injection in the baseline limits its ability to acquire precise spatial cues, while our approach leverages this mechanism to deliver both faster convergence and noticeably stronger performance.

In addition, introducing the gated mechanism incurs virtually no extra parameter overhead. As shown in Table~\ref{table:param}, the proposed gated control adds only 
0.09M parameters, which amounts to merely 
0.006\% of the original SANA model size, while delivering substantial improvements in both convergence speed and controllability.

\subsection{Ablation studies}
In addition to the analyses above, we also conduct extensive ablation studies on the proposed gated mechanism. Additional visualizations of the ablation studies are provided in the appendix.

\begin{table}[t]
    \centering
    \scriptsize
    \begin{tabular}{c|ccc}
        \toprule Method                    & FID $\downarrow$                       & SSIM $\uparrow$                                   & CLIP Score $\uparrow$                 \\
        \midrule \midrule 
        w/o gating &          22.6                      &      0.36                                      & 0.74                    \\
        w/o interaction & 20.0 & 0.40  & 0.76 \\
        After-FFN & \textbf{18.2} & 0.41 & \textbf{0.77} \\
        Elementwise & 18.8 & 
        \textbf{0.42}  & \textbf{0.77} \\
        Input features & 20.3 & 0.39  & 0.76\\
                                               
        \midrule
        Ours &\cellcolor{cvprblue!15} 19.0 & \cellcolor{cvprblue!15} \textbf{0.42} & \cellcolor{cvprblue!15} \textbf{0.77} \\
        \bottomrule
    \end{tabular}
    \caption{Ablation studies on the mechanism of
gated control for information integration.
    Best results are in \textbf{bold}.}
    \label{tab:ablation_studies}
\end{table}

\noindent\textbf{Whether to use gating.}
The convergence curves in Figure~\ref{fig:curve} clearly illustrate the benefits introduced by the gated mechanism, demonstrating both faster convergence and superior performance. In addition, Table~\ref{tab:ablation_studies} provides a more detailed quantitative comparison of evaluation metrics, further confirming the effectiveness of our approach. 
Removing the gated mechanism leads to a substantial degradation across multiple metrics, including FID, SSIM, and CLIP scores.
Experiments on the original OminiControl further demonstrate that our method also accelerates convergence on the softmax attention. Please refer to the appendix.
\vspace{1mm}

\noindent\textbf{Positions.}
We investigated three different insertion points for the gated mechanism: (a) after the self-attention layer, (b) after the cross-attention layer, and (c) after the Mix-FFN layer. Experiments show that applying the gated module after the cross-attention layer results in highly unstable training loss, whereas inserting it after the Mix-FFN layer leads to only minor performance fluctuations, as shown in Table~\ref{tab:ablation_studies}. 
We speculate that the cross-attention layer may impose stronger stability requirements on the mapped feature.
In practical sampling, we 
observe that our design produces outputs that are more consistent with human perception.

\vspace{1mm}
\noindent\textbf{Gated types.}
Different types of gated mechanisms are analyzed. As shown in Table~\ref{tab:ablation_studies}, applying element-wise gating achieves comparable performance, but at the cost of significantly more parameters (200M vs. 0.09M). In contrast, the tokenwise approach efficiently enables dynamic token selection and information fusion with minimal overhead. Directly summing the features, however, leads to unstable training loss, highlighting the critical role of dynamically selecting relevant token information.

\noindent\textbf{Input features.}
We also conduct a detailed study on the source of input features for the gated module to obtain token scores. We compare features taken before and after the self-attention layer, as shown in Table~\ref{tab:ablation_studies}, and find that using features from before the self-attention layer yields better performance. We hypothesize that this aligns with the token-wise scoring approach: by encouraging each token to independently predict its own score, it avoids potential interference from the gated layer’s gradients on normal attention interactions and enhances each token’s inherent ability to judge and process information autonomously.

\noindent\textbf{Whether to use interaction.}
Finally, we investigate the impact of omitting attention interactions. The results in Table~\ref{tab:ablation_studies} show that, for the Canny-to-image task, removing attention-based interactions negatively affects the model’s final performance, highlighting the importance of attention in token interaction.

%% file: tables/main_table.tex
\newcommand{\best}[1]{\textbf{#1}}

\begin{table*}
    [t] \scriptsize
    \centering
    \resizebox{\textwidth}{!}{
    \begin{tabular}{c|c|c|c|c|ccc|cc}
        \toprule \multirow{3}{*}{Task}           & \multirow{3}{*}{Methods / Setting} & \multirow{3}{*}{ Base Model} & \multirow{3}{*}{ Resolution} & Controllability                                & \multicolumn{3}{c|}{Image Quality}  & \multicolumn{2}{c}{Alignment}       \\
                                                 &        &                            &                              & \textbf{F1}$\uparrow$ / \textbf{MSE}$\downarrow$ & \textbf{FID}$\downarrow$            & \textbf{SSIM} $\uparrow$           & \textbf{MUSIQ}$\uparrow$                & \textbf{CLIP Text}$\uparrow$             & \textbf{CLIP Image}$\uparrow$       \\
        \midrule \midrule \multirow{4}{*}{Canny} & ControlNet                         & \multirow{2}{*}{SD1.5}    & \multirow{4}{*}{$512^2$}   & \textbf{0.35}                                           & \best{18.74}                        & \underline{0.36}                    & 67.81                                          & \textbf{0.305}                                    & \underline{0.752}                   \\
                                                 & T2I Adapter                        &                   &         & 0.22                                           & \underline{20.06}                & 0.35                                                    & 67.89                                                            & \best{0.305}                                    & 0.748                               \\
                                                 & OminiControl & \multirow{2}{*}{SANA} &  & 0.23 & 22.91 & 0.35 & \best{71.99} & \underline{0.263}	& 0.750 \\
                                                 & Ours &  &  & \underline{0.26}\cellcolor{cvprblue!15} & 21.97\cellcolor{cvprblue!15} & \best{0.37}\cellcolor{cvprblue!15} & \underline{71.86}\cellcolor{cvprblue!15} & \underline{0.263}\cellcolor{cvprblue!15}	& \best{0.762}\cellcolor{cvprblue!15} \\
        \midrule \multirow{4}{*}{Depth}          
        & ControlNet                 
        & \multirow{2}{*}{SD1.5}    & \multirow{4}{*}{$512^2$}   & 923                                            & \best{23.03}                        & \best{0.34}                                       & 70.73                                         & \underline{0.308}                                    & 0.726                   \\
                                                 & T2I Adapter                        &    &                          & 1560                                           & \underline{24.72}                               & \underline{0.28}                                  & 69.99                                                               & \best{0.309}                             & 0.721                               \\
                                  & OminiControl                      & \multirow{2}{*}{SANA}  &   & \underline{803} & 30.95 & \best{0.34} & \underline{71.65} & 0.265 & \underline{0.735}  \\
                                                 & Ours                              &     &                         & \best{626}\cellcolor{cvprblue!15}              & 30.15\cellcolor{cvprblue!15}        & \best{0.34}\cellcolor{cvprblue!15} & \best{72.30}\cellcolor{cvprblue!15} & 0.265\cellcolor{cvprblue!15}             & \best{0.739}\cellcolor{cvprblue!15} \\
                                                 \midrule 
                               \multirow{3}{*}{Deblurring}   & ControlNet                   & \multirow{3}{*}{SANA}   & \multirow{3}{*}{$1024^2$}   & 128                                            & 69.24                              & 0.58                                                 & 43.29                                           & 0.222                                   & 0.723                          \\
                                             & OminiControl & & & \underline{120} & \underline{10.65} & \underline{0.61} & \best{67.28} & \best{0.253} & \underline{0.896} \\
                                                 & Ours          &                     &                              & \best{14}\cellcolor{cvprblue!15}               & \best{7.45}\cellcolor{cvprblue!15} & \best{0.64}\cellcolor{cvprblue!15} & \underline{65.59}\cellcolor{cvprblue!15}       & \underline{0.252}\cellcolor{cvprblue!15}      & \best{0.934}\cellcolor{cvprblue!15} \\
        \midrule \multirow{2}{*}{Colorization}           & ControlNet                         & \multirow{2}{*}{SANA}   & \multirow{2}{*}{$1024^2$}
        & 171                                           & 24.95                               & \textbf{0.80}                                                   & 47.39                                               & 0.239                                   & 0.842                              \\
                                                                           & Ours                               &                    &          & \best{163}\cellcolor{cvprblue!15}             & \best{10.28 }\cellcolor{cvprblue!15} & 0.79\cellcolor{cvprblue!15} & \best{54.85}\cellcolor{cvprblue!15}          & \best{0.254   }\cellcolor{cvprblue!15}      & \best{0.897 }\cellcolor{cvprblue!15} \\
        \midrule \multirow{2}{*}{HED}   & ControlNet                   & \multirow{2}{*}{SANA}    & \multirow{2}{*}{$1024^2$}  & 2320                                            & 20.36                              & 0.39                                               & 55.58                                 & 0.238                                & 0.733                       \\
                                                 & Ours     &                          &                              & \best{1168}\cellcolor{cvprblue!15}               & \best{16.81}\cellcolor{cvprblue!15} & \best{0.47}\cellcolor{cvprblue!15} & \best{63.06}\cellcolor{cvprblue!15}       & \best{0.259}\cellcolor{cvprblue!15}      & \best{0.798}\cellcolor{cvprblue!15} \\
        \midrule
    \end{tabular}
    } 
    \vspace{-0.4cm}
    \caption{Quantitative comparison with baseline methods on five spatially
    aligned tasks. We evaluate methods based on Controllability (F1-Score for
    Canny, MSE for others), Image Quality (SSIM, FID and MUSIQ),
    and Alignment (CLIP Text and CLIP Image). For F1-Score (used in Canny to Image
    task), higher is better; for MSE, lower is better. Best results are shown in
    \best{bold}. The second-best results are highlighted with underlines.}
    \vspace{-1em}
    \label{tab:main_results}
\end{table*}

%% file: sec/5_conclusion.tex
\section{Conclusion}
We propose a unified gated framework for controllable diffusion on linear-attention backbones, addressing the limitations of ControlNet and OminiControl. Our gating mechanism enables stable and expressive multi-type conditioning, achieving superior controllability and faster convergence with minimal parameter overhead. Extensive experiments on spatially aligned and subject-driven tasks demonstrate high-quality generation and robust control across diverse conditions, highlighting the potential of efficient, on-device controllable diffusion.

%% file: sec/acknowledge.tex
\section*{Acknowledgements}
This project is supported by the Ministry of Education, Singapore, under its Academic Research Fund Tier 2 (Award Number: MOE-T2EP20122-0006),
the National Research Foundation, Singapore, and Cyber Security
Agency of Singapore under its National Cybersecurity R\&D Programme and CyberSG R\&D Cyber
Research Programme Office (Award: CRPO-GC1-NTU-002).

%% file: sec/X_suppl.tex
\clearpage
\setcounter{page}{1}
\maketitlesupplementary

\appendix
\setcounter{figure}{0}
\setcounter{table}{0}

\renewcommand{\thefigure}{\arabic{figure}}
\renewcommand{\thetable}{\arabic{table}}

\section{Gated control on OminiControl}
\begin{figure}
    \centering
    \includegraphics[width=0.95\linewidth]{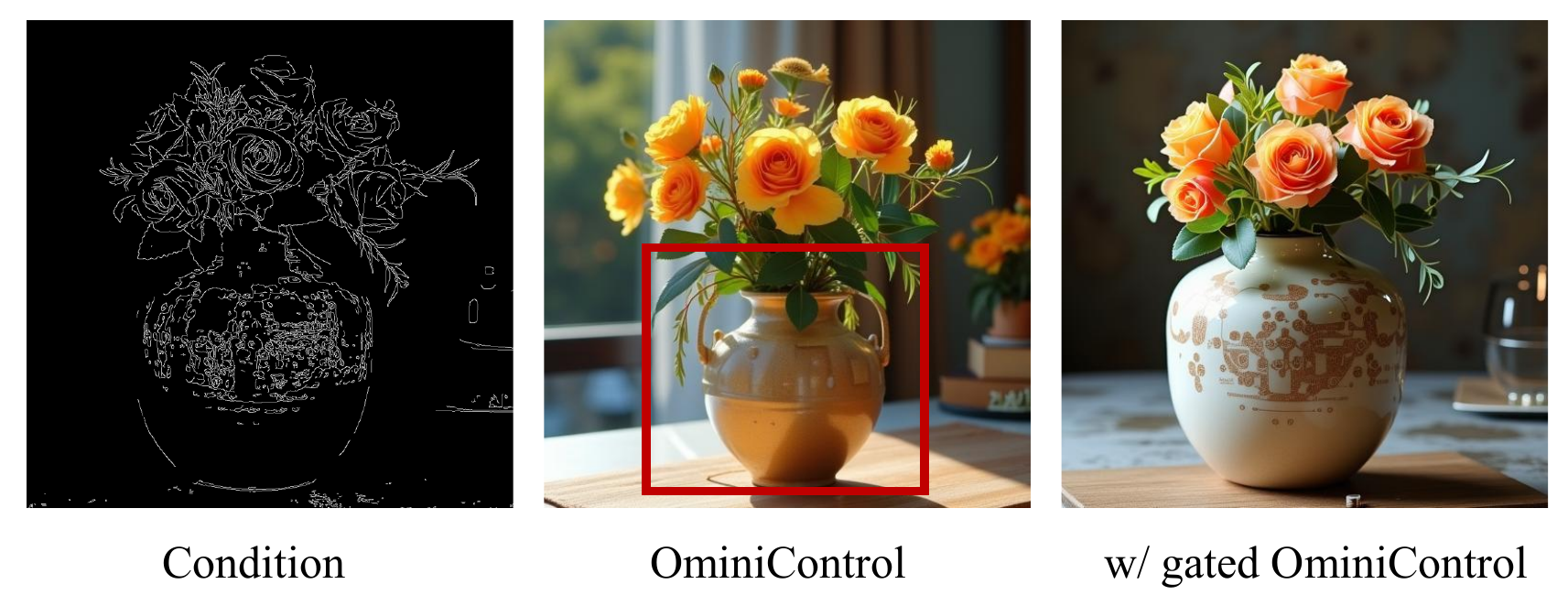}
    \caption{Gated control on the original OminiControl. Our approach enables the model to capture control signals much earlier during training, even when using softmax attention.}
    \label{fig:omini}
\end{figure}

To further validate the generality of our gated mechanism on the softmax attention, we implement a minimal variant on top of the original OminiControl~(based on the FLUX.1-dev) by introducing a single-block gated interaction between image-condition tokens and latent tokens, adding only 0.2M parameters. 
Even with this lightweight modification, our approach enables the model to grasp control information much earlier during training.
Figure~\ref{fig:omini} presents the test-time outputs at step 400 under strictly identical training configurations. With the addition of a lightweight gated control mechanism, the model achieves significantly faster alignment with the vase specified in the condition input.

\begin{figure}
    \centering
    \includegraphics[width=0.99\linewidth]{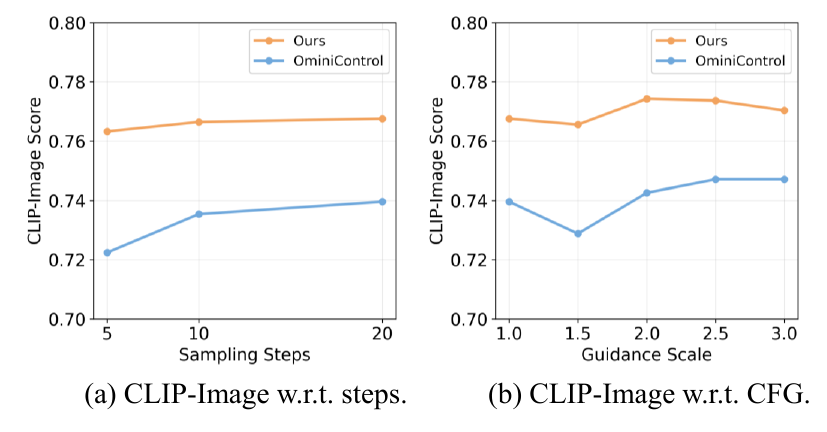}
    \vspace{-2mm}
    \caption{Robustness to sampling steps and guidance scale. Our model produces better and more stable outputs than OminiControl under both low-step inference and varying guidance scales.}
    \label{fig:robust}
\end{figure}

\begin{figure}
    \centering
    \includegraphics[width=0.92\linewidth]{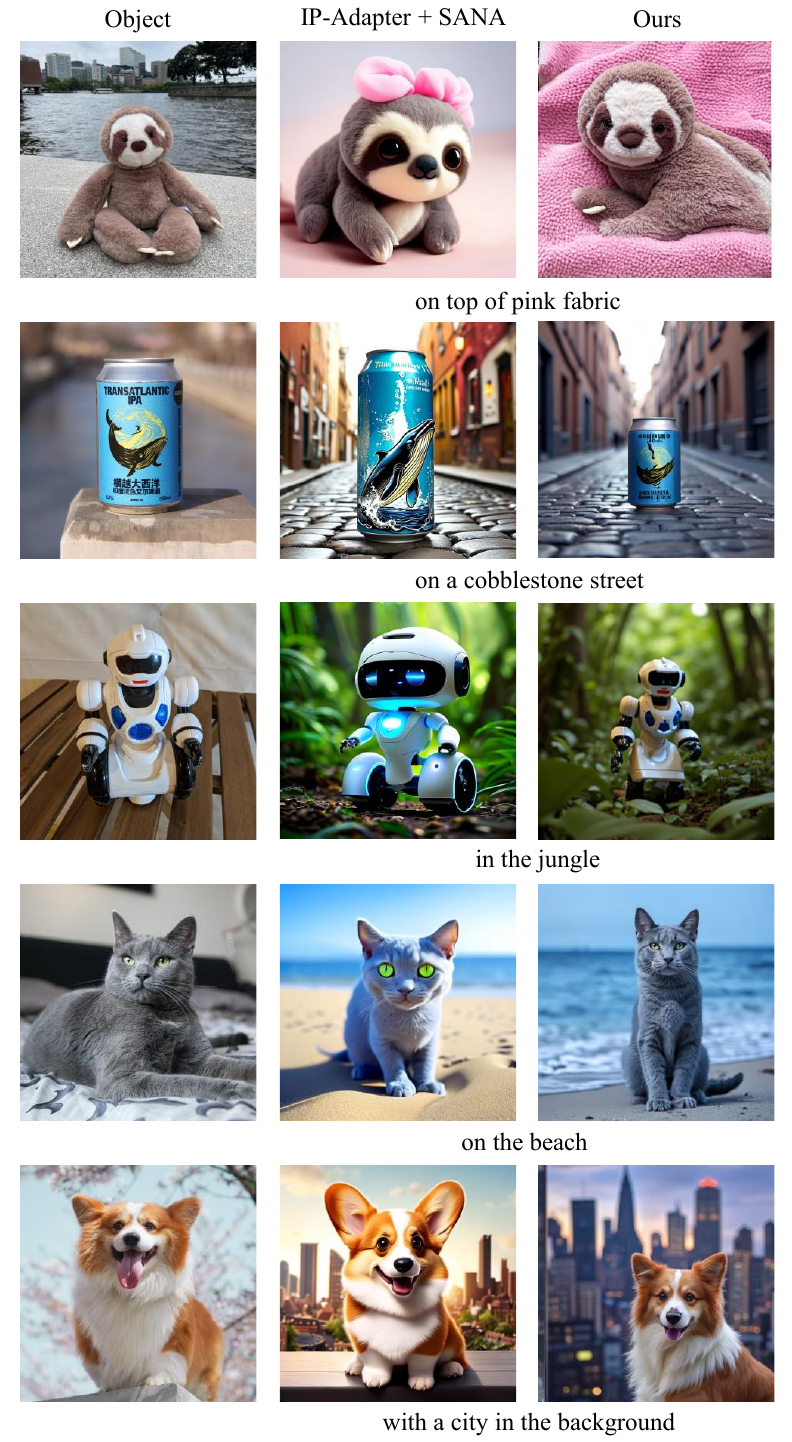}
    \vspace{-2mm}
    \caption{Further visual comparisons on the subject-driven tasks. Our approach preserves object-specific features with greater fidelity, while simultaneously adapting the environment according to the provided editing prompt in a natural and flexible manner.}
    \label{fig:subject1}
\end{figure}

\begin{table*}
    [t]
    \centering
    \begin{tabular}{l|ccccc|c}
        \toprule Method                                                  & Identity      & Material      & Color         & Natural        & Modification  & Average       \\
                                                                         & preservation  & quality       & fidelity      & appearance     & accuracy      & score         \\
        \midrule
        IP-Adapter (SANA)                                                & 24.8          & 30.4          & 37.2          & 56.1           & 44.8          & 38.7          \\
        \rowcolor{cvprblue!15} Ours                                      & \textbf{52.9} & \textbf{63.5}          & \textbf{58.4} & \textbf{72.4}  & \textbf{55.6} & \textbf{60.6} \\
        \bottomrule
    \end{tabular}
    \caption{Quantitative evaluation results (in percentage) across different
    evaluation criteria. Higher values indicate better performance.}
    \label{tab:quant_results}
\end{table*}

\begin{figure*}[htbp] 
    \centering
    \includegraphics[width=0.7\linewidth]{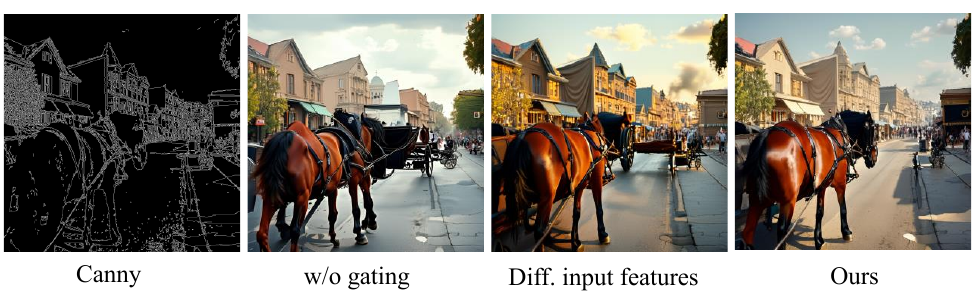}
    \caption{Intuitive analysis of the ablation on gated application through visualizations. Without gating, the model’s ability to leverage conditional information is substantially impaired, which may lead to outputs that do not comply with the condition or generate unreasonable artifacts. Moreover, using input features from different sources to generate the gate scores can affect global aspects of the image, such as style and color.}
    \label{fig:ablation}
\end{figure*}

\section{Visual analysis of ablations}
We further provide a more intuitive analysis of the ablation on gated application through visualizations, as shown in Figure~\ref{fig:ablation}. Without gating, the model’s ability to leverage conditional information is substantially impaired, which may lead to outputs that do not comply with the condition or generate unreasonable artifacts, such as unnatural protrusions on a horse’s back, and a decrease in overall image quality. Moreover, using input features from different sources to generate the gate scores can affect global aspects of the image, such as style and color. This observation suggests that during the model’s forward pass, input features from different positions may exhibit a certain degree of semantic progression, for example, a growing sensitivity to style and color.

\begin{figure}
    \centering
    \includegraphics[width=0.99\linewidth]{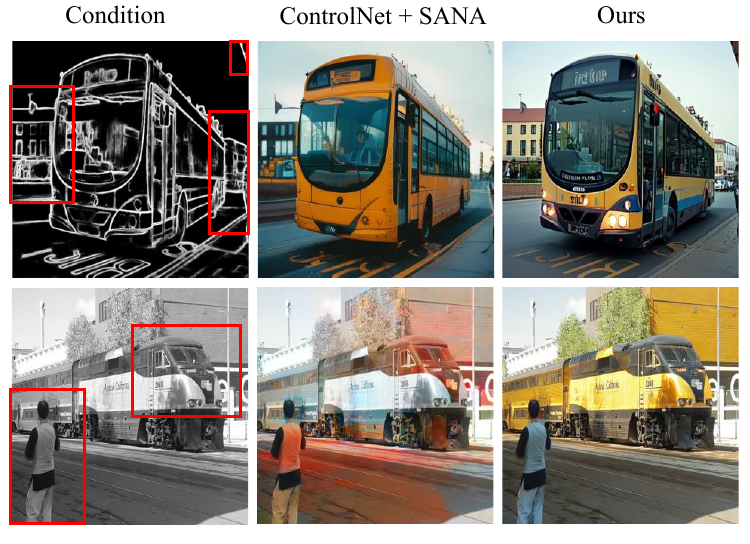}
    \caption{Comparison between our approach and SANA-based ControlNet. Our method more accurately follows the conditions and can generate more natural and realistic colorization.}
    \label{fig:bus}
\end{figure}

\section{Sampling steps and guidance scale}
We further evaluate the robustness of our model with respect to sampling steps and guidance scale.
The results in Figure~\ref{fig:robust}
indicate that our model produces \textbf{better} and more \textbf{stable} outputs than OminiControl under both low-step inference and varying guidance scales, further supporting the suitability for low-latency scenarios.

\section{Image editing and multi-condition} 
\begin{figure}
    \centering
    \includegraphics[width=0.99\linewidth]{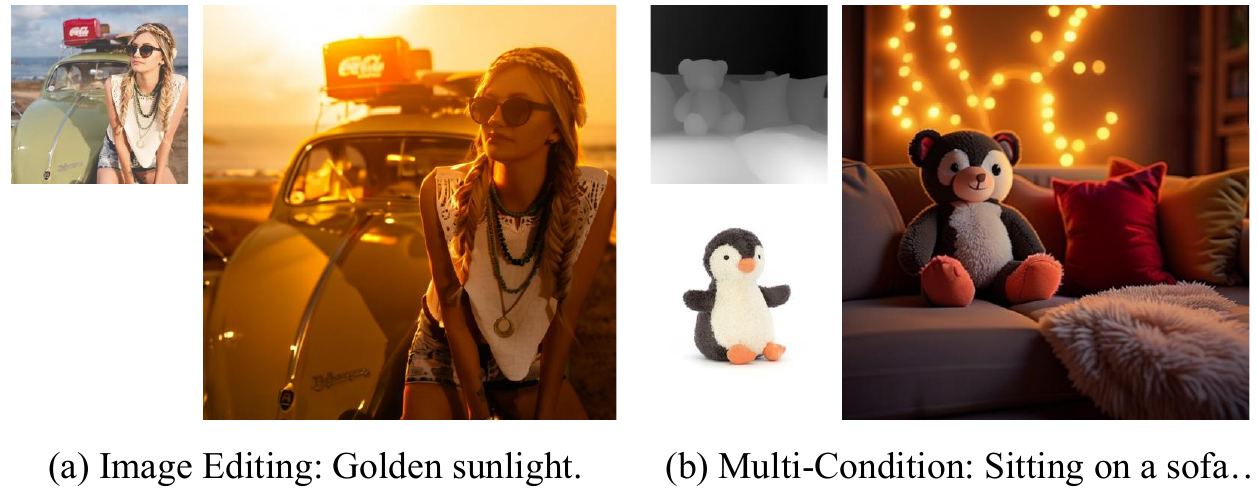}
    \caption{Examples of image editing and multi-condition control. Our model can acquire basic editing capability within limited training steps. Moreover, it is able to simultaneously incorporate multiple conditions. }
    \label{fig:editing}
\end{figure}

GateControl can acquire basic editing capability within limited training steps, as shown in Figure~\ref{fig:editing}(a), demonstrating fast adaptation and broad applicability. We further conduct multi-condition experiments (subject + depth), as depicted in Figure~\ref{fig:editing}(b), showing that our model can simultaneously incorporate multiple conditions. However, when multiple conditions are combined, conflicts may arise; for example, satisfying geometric constraints may slightly alter the original subject shape.

\section{Comparison on the subject-driven tasks}

\subsection{Evaluation for subject-driven generation}
\label{subsec:ap:sub} \textbf{Evaluation criteria.} 
Following OminiControl, we utilize a five-dimensional evaluation protocol that systematically measures both fidelity to subject characteristics and compliance with user-specified modifications. The assessment dimensions comprise Identity Preservation, Material Quality, Color Fidelity, Natural Appearance, and Modification Accuracy. Evaluations are conducted using the GPT-4o multimodal model.

\noindent\textbf{Results comparison.} Results compared with the SANA-based IP-Adapter are presented in Table~\ref{tab:quant_results}.
These results demonstrate that our method substantially surpasses IP-Adapter on subject-driven generation. It achieves markedly stronger preservation of object-specific attributes, provides more flexible and faithful control over user-specified modifications, and produces outputs that exhibit more natural integration with the surrounding context.

\subsection{Further visual comparisons}
Figure~\ref{fig:subject1} illustrates the qualitative differences between our method and the SANA-based IP-Adapter. Our approach preserves object-specific features with greater fidelity, while simultaneously adapting the environment according to the provided editing prompt in a natural and flexible manner, significantly surpassing the IP-Adapter results.
Furthermore, our approach is generalizable: the same architecture can handle both subject-driven and spatially-aligned tasks, with only LoRA and gating parameters adjusted.

\section{Comparisons on spatially aligned tasks}
Figure~\ref{fig:bus} presents a comparison between our approach and SANA-based ControlNet. Our method more accurately follows the conditions, generating surrounding objects like vehicles and buildings more faithfully, and produces higher-quality outputs for the bus. For the coloring task, our model achieves more natural and realistic colorization.